\documentclass{article}

     \PassOptionsToPackage{numbers, compress}{natbib}

     \usepackage[preprint]{neurips_2019}

\usepackage[utf8]{inputenc} 
\usepackage[T1]{fontenc}    
\usepackage{hyperref}       
\usepackage{url}            
\usepackage{booktabs}       
\usepackage{amsfonts}       
\usepackage{nicefrac}       
\usepackage{microtype}      
\usepackage{epsfig}
\usepackage{graphicx}
\usepackage{subfigure} 
\usepackage{floatrow}
\title{GAPNet: Graph Attention based Point Neural Network for Exploiting Local Feature of  Point Cloud}

\author{%
  Can Chen \\
  School of SATM \\
  Cranfield University \\
  UK, MK43 0AL \\
  \texttt{can.chen@cranfield.ac.uk} \\
\And
  Luca Zanotti Fragonara \\
  School of SATM \\
  Cranfield University \\
  UK, MK43 0AL \\
  \texttt{l.zanottifragonara@cranfield.ac.uk} \\
\AND
  Antonios Tsourdos \\
  School of SATM \\
  Cranfield University \\
  UK, MK43 0AL \\
  \texttt{a.tsourdos@cranfield.ac.uk} \\
}

\begin{document}

\maketitle

\begin{abstract}
  Exploiting fine-grained semantic features on point cloud is still challenging due to its irregular and sparse structure in a non-Euclidean space. Among existing studies, PointNet provides an efficient and promising approach to learn shape features directly on unordered 3D point cloud and has achieved competitive performance. However, local feature that is helpful towards better contextual learning is not considered.  Meanwhile, attention mechanism shows efficiency in capturing node representation on graph-based data by attending over neighboring nodes. In this paper, we propose a novel neural network for point cloud, dubbed GAPNet, to learn local geometric representations by embedding graph attention mechanism within stacked Multi-Layer-Perceptron (MLP) layers. Firstly, we introduce a GAPLayer to learn attention features for each point by highlighting different attention weights on neighborhood. Secondly, in order to exploit sufficient features, a multi-head mechanism is employed to allow GAPLayer to aggregate different features from independent heads. Thirdly, we propose an attention pooling layer over neighbors to capture local signature aimed at enhancing network robustness. Finally, GAPNet applies stacked MLP layers to attention features and local signature to fully extract local geometric structures. The proposed GAPNet architecture is tested on the ModelNet40 and ShapeNet part datasets, and achieves state-of-the-art performance in both shape classification and part segmentation tasks.
\end{abstract}

\section{Introduction}

As point cloud data becomes increasingly popular in a wide range of applications such as: autonomous vehicle \cite{zhou2018voxelnet, qi2018frustum, ku2018joint, liu2018real}, robotic mapping and navigation \cite{biswas2012depth, zhu2017target}, 3D shape representation and modelling \cite{golovinskiy2009shape}, many researchers are drawing attention to shape analysis and understanding, especially when convolutional neural networks (CNNs) achieves significant success in computer vision tasks. However, CNNs heavily rely on the data with the standard grid structure, which leads to inefficient performance on irregular and unordered geometric data, such as point cloud. As a result, fully exploiting contextual information from point cloud remains a challenging problem. 

In order to leverage advantages of CNNs, some approaches \cite{maturana2015voxnet, wang2015voting, riegler2017octnet} map unstructured point cloud to a standard 3D grid before applying CNN architectures. However, these volumetric representations are not efficient in terms of memory and computational efficiency due to the typical sparsity of point cloud structure. Instead of applying CNNs over gridded point cloud, \textit{PointNet} \cite{qi2017pointnet} pioneers the approach that applies deep learning directly over irregular point cloud. In particular, \textit{PointNet} makes input point cloud invariant to permutations and exploits point-wise features by independently applying a Multi-Layer-Perceptron (MLP) network and a symmetric function on each point. However, it only captures global feature without local information. \textit{PointNet++} \cite{qi2017pointnet++} extends \textit{PointNet} model by constructing a hierarchical neural network that recursively applies \textit{PointNet} with designed sampling and grouping layers to extract local features. \textit{DGCNN} \cite{wang2018dynamic} operates an edge convolution on points and corresponding edges to further exploit local information. Adapted from point cloud registration method, \textit{KC-Net} \cite{shen2018mining} builds a kernel correlation layer to measure geometric affinities for points.

Attention mechanisms have proved to be efficient in many areas, such as machine translation task \cite{vaswani2017attention, bahdanau2014neural}, vision-based task \cite{mnih2014recurrent}, and graph-based task \cite{velivckovic2017graph}. Inspired by graph attention networks \cite{velivckovic2017graph}, we primarily focus on fully exploiting fine-grained local features for point cloud in an attention manner in 3D shape classification and part segmentation tasks. The key contributions of our work are summarized as follows:

\begin{itemize}
\item We propose a multi-head GAPLayer to capture contextual attention features by indicating different \textit{importance} of neighbors for each point. Independent heads attend to different features from representation space in parallel and are further aggregated together to obtain sufficient power of feature extraction.
\item We propose self-attention and neighboring-attention mechanisms to allow the GAPLayer to compute the attention coefficients by considering the self-geometric information and local correlations to corresponding neighbors.
\item An attention pooling layer over neighbors is proposed to identify the most important features to obtain local signature representation to enhance network robustness.
\item Our GAPNet integrates the GAPLayer and the attention pooling layer into stacked Multi-Layer-Perceptron (MLP) layers or existing pipelines (e.g. \textit{PointNet}) to better extract local contextual feature from unordered point cloud.
\end{itemize}

\section{Related work}

\paragraph{Learning features from volumetric grid.} Voxelization is an intuitive way to convert sparse and irregular point cloud to standard grid structure, after which standard CNNs can be applied for feature extraction. \textit{Voxnet} \cite{maturana2015voxnet} voxelizes the point cloud into a volumetric grid that indicates spatial occupancy for each voxel, followed by a 3D CNN over occupied voxels to predict categories of objects. However, 3D dense and sparsely-occupied volumetric grid leads to large memory and computational cost for high spatial resolution. As a result, some improvements are proposed to address the sparsity problem. \textit{Kd-Net} \cite{klokov2017escape} uses a kd-tree \cite{bentley1975multidimensional} to build an efficient 3D space partition structure and a deep architecture to learn representations of point cloud. Similarly, \textit{OctNet} \cite{riegler2017octnet} applies 3D convolution on a hybrid grid-octree structure generated from a set of shallow octrees to achieve high resolution.

\paragraph{Learning features from unstructured point cloud directly.} \textit{PointNet} \cite{qi2017pointnet} is the pioneer work that proposed the direct application of deep learning on the raw point cloud. In more detail, a Multi-Layer-Perceptron (MLP) network and a symmetric function (e.g. max pooling) are applied on every individual point to extract global feature. This approach provides an efficient way for unstructured point cloud understanding, however, local feature is not captured as the architecture only works on independent points without relationships measurement between points in the local regions. To address this problem, PointNet++ \cite{qi2017pointnet++} constructs a hierarchical neural network that recursively applies \textit{PointNet} with a sampling layer and a grouping layer to exploit local representations. \textit{DGCNN} \cite{wang2018dynamic} extends \textit{PointNet} by presenting an edge convolution operation (EdgeConv) that is applied on edge features which aggregate each point and corresponding edges connecting to the neighboring pairs. In order to leverage the advantages of standard CNN operation, \textit{PointCNN} \cite{li2018pointcnn} attempts to learn a \(\chi\)-convolutional operator to transform a given unordered point set to a latent canonical order, after which a typical CNN architecture is used to extract local features.

\paragraph{Learning features from multi-view models.} In order to apply standard CNN operation but also avoid large computation cost in volumetric-based methods, some researchers are interested in multi-view based approaches. For instance, \cite{qi2016volumetric, wang2017dominant} learns features of point cloud in an indirect way by applying a typical 2D CNN architecture to multiple 2D image views that are generated by the multi-view projections over 3D point cloud. However, these multi-view approaches are not capable to realize semantic segmentation task for point cloud, as 2D images lack of depth information, which leads to the fact that it is non-trivial to classify each point from images.

\paragraph{Learning features from geometric deep learning.} Geometric deep learning \cite{bronstein2017geometric} is a modern term for a set of emerging techniques that attempts to address non-Euclidean structured data (e.g. 3D point cloud, social networks or genetic networks) by deep neural networks. Graph CNNs \cite{bruna2013spectral, defferrard2016convolutional, zhang2018graph} show advantages of graph representation in many tasks for non-Euclidean data, as it can naturally deal with these irregular structures. \textit{PointGCN} \cite{ZhangR_18_gcnn_point_cloud} builds a graph CNN architecture to capture local structure and classify point cloud, which also proves that geometric deep learning has huge potential for unordered point cloud analysis.

\section{GAPNet architecture}
In this section, we propose our GAPNet model to better learn local representations for unstructured point cloud in shape classification and part segmentation tasks. We detail the model that consists of three components: GAPLayer (multi-head graph attention based point network layer) that is shown in Figure \ref{fig:attention_structure} ,  attention pooling layer, and GAPNet architecture shown in Figure \ref{fig:model_structure} .

Let \(X=\left\{ x_i \in \mathbb{R}^F, i=1,2,\ldots,N\right\}\) be a raw set of unordered points and input of our model, with \(F\)-dimension, where \(N\) is the number of the points, and \(x_i\) is a feature vector that might contain 3D space coordinates \((x,y,z)\), color, intensity, surface normal, etc. For the sake of simplicity, in this study we set \(F=3\) and only use 3D coordinates as input features.

\subsection{GAPLayer}
\paragraph{Local structure representation.} Considering the fact that the number of samples in point cloud can be very large in real applications (e.g. autonomous vehicle), allowing every point to attend to all other points will lead to high computation cost and gradient vanishing problem due to very small weights allocation on every other point for every point. As a result, we construct a directed \(k\)-nearest neighbor graph \(G=(V,E)\) to represent local structure of the point cloud, where \(V=\left\{1,2,\ldots,N\right\}\) are nodes for points, \(E \subseteq V \times N_i\) are edges connecting neighboring pairs of points, and \(N_i\) is a neighborhood set of point \(x_i\). We define edge features as  \({y_i}_j=(x_i-{x_{i}}_j)\), where \(i \in V, j \in N_i\), and \({x_i}_j\) indicates the neighboring point \(x_j\) to point \(x_i\).

\begin{figure}[h]
  \centering
   {\epsfig{file = 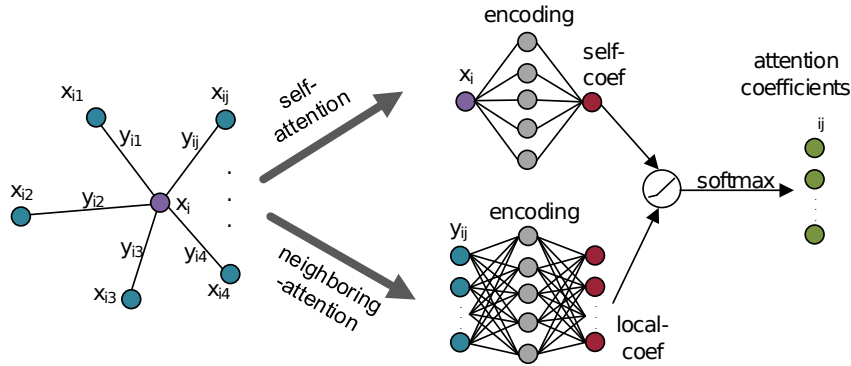, width=0.8\linewidth}}
  \caption{An illustration of attention coefficients generation. \(x_i\) and \({x_{i}}_j\) denote a certain point and corresponding neighbors respectively, and \({y_i}_j\) are corresponding edges. Self-coefficients (self-coef for short) and local-coefficients (local-coef for short) are fused by a leaky RELU activation function, and further normalized by softmax function to generate attention coefficients for neighboring pairs.}
  \label{fig:attention}
 \end{figure}

\begin{figure*}[t!]
  \centering
   \subfigure[GAPLayer]{\label{fig:multi_attn}\includegraphics[width=0.4\linewidth]{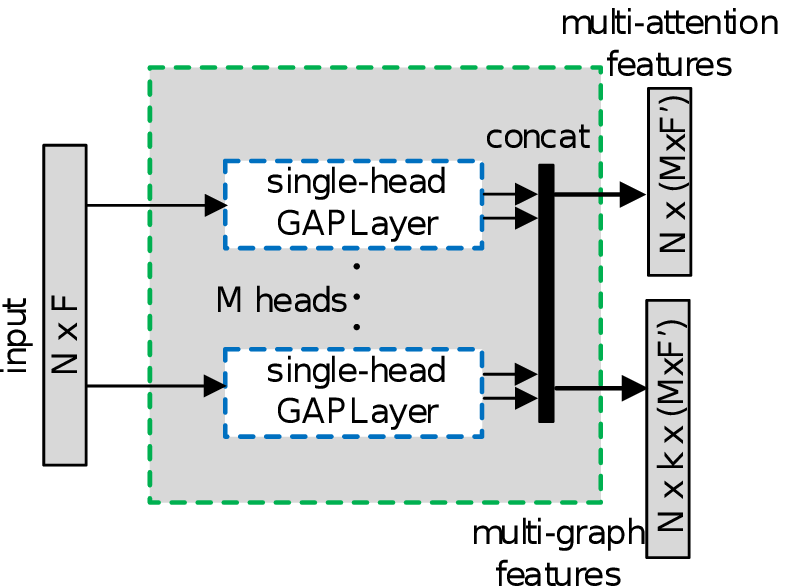}}
   \subfigure[Single-head GAPLayer]{\label{fig:single_attn}\includegraphics[width=0.55\linewidth]{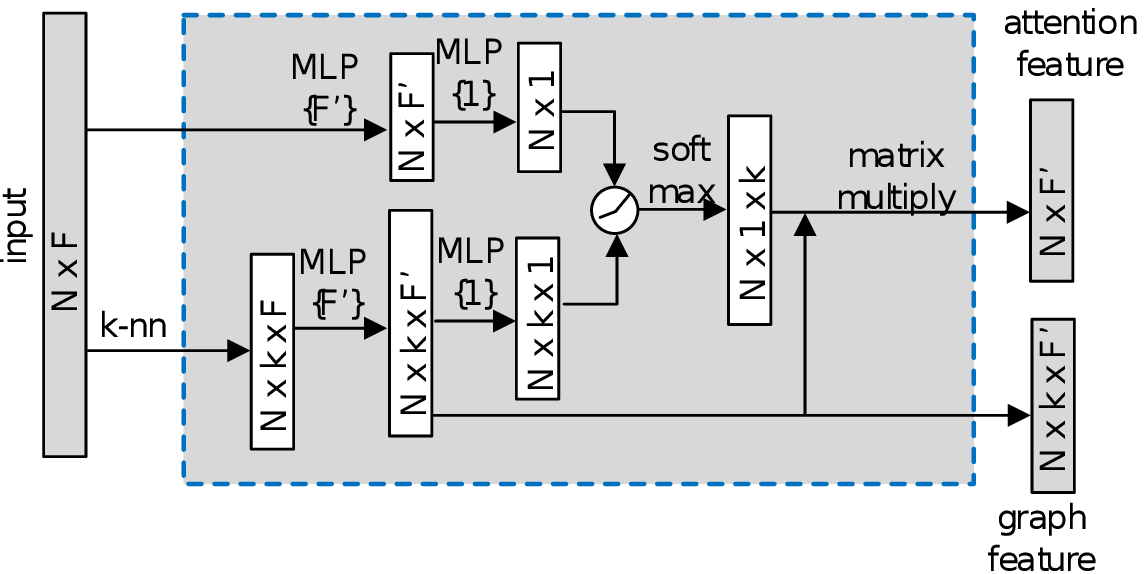}}
  \caption{\textbf{GAPLayer structure. } The GAPLayer with \(M\) heads,  as shown in ~\ref{fig:multi_attn} , takes \(N\) points with \(F\) dimensions as input and concatenates attention feature and graph feature respectively from all heads to generate multi-attention features and multi-graph features as output. As shown in ~\ref{fig:single_attn} , the single-head GAPLayer learns self-attention and neighboring-attention features in parallel that are then fused together by a non-linear activation function leaky RELU to obtain attention coefficients, which are further normalized by a softmax function, then a linear combination operation is applied to finally generate attention feature.  MLP\{\} denotes multi-layer perceptron operation, numbers in brace stand for size of a set of filters, and we use the same notation for the remainder.}
  \label{fig:attention_structure}
 \end{figure*}

\paragraph{Single-head GAPLayer.} To the benefit of the readers, we start by introducing a single-head GAPLayer that takes point cloud data as the input, jointly with a multi-head mechanism that concatenates all heads together over feature channels in our network. The structure of single-head GAPLayer is shown in Figure \ref{fig:single_attn} .

In order to pay different attentions to different neighbors, we propose a self-attention mechanism and a neighboring-attention mechanism to capture attention coefficients for each point to its neighborhood as illustrated in Figure \ref{fig:attention} . In more detail, the self-attention mechanism learns self-coefficients by considering self-geometric information for each individual point, while neighboring-attention mechanism focuses on local-coefficients by considering neighborhood.

As an initial step, we encode nodes and edges of point cloud with respect to to the higher-level features with output dimension \(F'\) as defined by Equation \ref{eq:new_feature} and \ref{eq:new_edge} .

\begin{equation}\label{eq:new_feature}
    x'_i=h(x_i,\theta)
\end{equation}

\begin{equation}\label{eq:new_edge}
    {y'_i}_j=h({y_i}_j, \theta)
\end{equation}

where h() is a parametric non-linear function, chosen to be a single-layer neural network in our experiment, and \(\theta\) is a set of learnable parameters of the filter.

We obtain attention coefficients by fusing self-coefficients  \(h(x'_i,\theta)\) and local-coefficients \(h({y'_i}_j,\theta)\) as defined by Equation \ref{eq:precoef} , where \(h(x'_i,\theta)\) and \(h({y'_i}_j,\theta)\) are single-layer neural network with 1-dimension output. \(LeakyReLU()\) denotes non-linear activation function leaky RELU.

\begin{equation}\label{eq:precoef}
    {c_i}_j=LeakyReLU(h(x'_i,\theta)+h({y'_i}_j,\theta))
\end{equation}

In order to align comparison of the attention coefficients across neighbors for different points, we use softmax function to normalize coefficients for all the neighbors to every point that is referred as \ref{eq:coef} .

\begin{equation}\label{eq:coef}
    {\alpha_i}_j=\frac{exp({c_i}_j)}{\sum_{k\in N_i}exp({c_i}_k)}
\end{equation}

The goal of each single-head GAPLayer is to compute contextual attention feature for every point. For this, we utilize the obtained normalized coefficients to compute a linear combination that is shown in Equation \ref{eq:attentional_node} . As shown in Figure \ref{fig:single_attn} , the outputs of single-head GAPLayer are attention feature \(\hat{x}_i \in \mathbb{R}^{F'}\) and graph feature encoded from graph edges.

\begin{equation}\label{eq:attentional_node}
    \hat{x}_i=f(\sum_{j \in N_i}{\alpha_i}_j{y'_i}_j)
\end{equation}

Where f() is a non-linear activation function, chosen to be RELU in our experiment.

\paragraph{Multi-head mechanism.} In order to obtain sufficient structural information and stabilize the network, we concatenate \(M\) independent single-head GAPLayers to generate a multi-attention features with \(M \times F'\) channels. The equation is defined as \ref{eq:multi_head}  . As shown in Figure \ref{fig:multi_attn} , the outputs of multi-head GAPLayer (GAPLayer for short) are multi-attention features and multi-graph features that concatenate attention feature and graph feature respectively from corresponding head.

\begin{equation}\label{eq:multi_head}
    \hat{x}'_i=\mathop{\Vert}\limits_{m}^M\hat{x}_i^{(m)}
\end{equation}

Such that \(\hat{x}_i^{(m)}\) is the attention feature of the \(m\)-th head, \(M\) is the total number of heads, and \(\parallel\) is concatenation operation over feature channels.

\subsection{Attention pooling layer}
To enhance network robustness and improve performance, we define an attention pooling layer on neighboring channel of multi-graph features. We use max pooling as our attention pooling operation which identifies the most important feature across heads to capture local signature representation \(Y_i\) defined as \ref{eq:maxpooling}  . The local signature is connected to the intermediate layer for capturing global feature.

\begin{equation}\label{eq:maxpooling}
    Y_i=\mathop{\Vert}\limits_{m}^M\mathop{\max}\limits_{j \in N_i}{y'_i}_j^{(m)}
\end{equation}

\begin{figure}[t!]
  \centering
   {\epsfig{file = 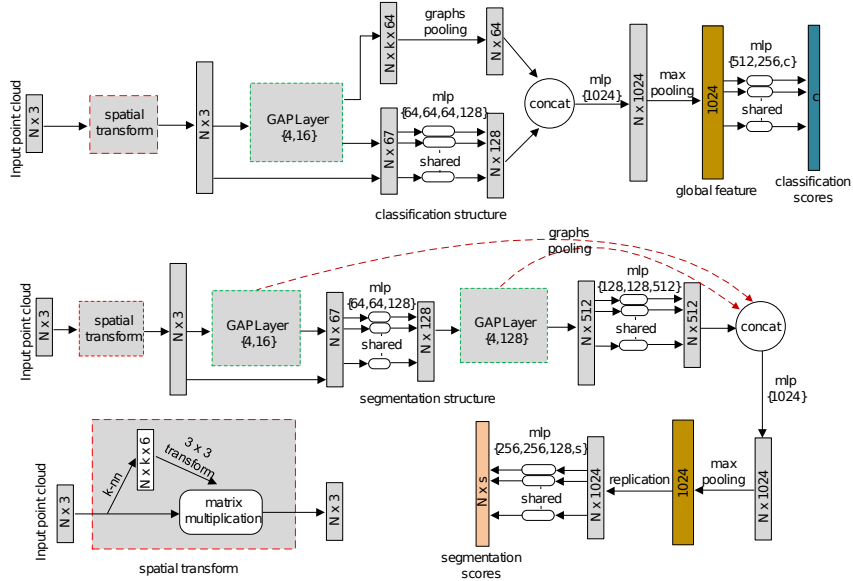, width=0.8\linewidth}}
  \caption{\textbf{GAPNet architecture:} The architecture contains two parts: classification (top branch) and semantic part segmentation (bottom branch). The classification model takes \(N\) points as input and applies one GAPLayer to obtain multi-attention features and multi-graph features that then apply shared MLP layers and attention pooling layer respectively, followed by a shared full-connected layer to form a global feature that is used to obtain classification scores for \(c\) categories. The semantic part segmentation model (bottom branch) extends the classification model by a second GAPLayer with MLP layers to obtain a certain of part category for each point from \(s\) semantic labels. Two red arrowed arcs represent attention pooling operation from corresponding GAPLayer to generate local signature that is concatenated to intermediate layer for global feature generation. Besides, GAPLayer\{4,16\} denotes a GAPLayer with 4 heads and 16 channels of encoding feature. \textbf{spatial transform network:} The spatial transform network is used to make point cloud invariant to certain transformations. The model learns a \(3 \times 3\) matrix for affine transformation from a single-head GAPLayer with 16 channels.}
  \label{fig:model_structure}
 \end{figure}

\subsection{GAPNet architecture}
Our GAPNet model shown in Figure \ref{fig:model_structure} considers both shape classification and semantic part segmentation for point cloud. The architecture is similar to \textit{PointNet} \cite{qi2017pointnet}. However, there are three main differences between the architectures. Firstly, we use an attention-aware spatial transform network to make the point cloud invariant to certain transformations. Secondly, instead of only processing individual points, we exploit local features by a GAPLayer before the stacked MLP layers. Thirdly, an attention pooling layer is used to obtain local signature that is connected to the intermediate layer for capturing a global descriptor.

\section{Experiments}
In this section, we evaluate our GAPNet model in the classification and part segmentation tasks for 3D point cloud analysis, we then compare our performance to recent state-of-the-art methods and perform ablation study to investigate different design variations.

\subsection{Classification}
\label{cls_point}
\paragraph{Dataset.} We demonstrate the effectiveness of our classification model on the ModelNet40 benchmark \cite{wu20153d} for shape classification. The ModelNet40 dataset contains 12,311 meshed CAD models that are classified to 40 man-made categories. We separate 9,843 models for training and 2,468 models for testing. Then we normalize the models in the unit sphere and uniformly sample 1,024 points over model surface. Besides, We further augment the training dataset by randomly rotating, scaling the point cloud and jittering the location of every point by means of Gaussian noise with zero mean and 0.01 standard deviation for all the models.

\paragraph{Network structure.} The classification model is presented in Figure \ref{fig:model_structure}  (top branch). In order to make the input points invariant to some geometric transformations, such as scale, rotation, we firstly apply an attention-aware spatial transformer network to align the point cloud to a canonical space. The network employs a single-head GAPLayer with 16 channels to capture attention features, followed by three shared MLP layers (64, 128, 1024) to output neurons with sizes 64, 128, 1024 respectively, then a max pooling operation and two full-connected layers (512, 256) are used to finally generate a transformation matrix.

A multi-head GAPLayer is then applied to generate multi-attention features with \(M \times F'\) channels, where the number of heads is set as \(M=4\), and the number of encoding channels is set as \(F'=16\). Our multi-attention features aggregate coordinate feature of point cloud to obtain a contextual attention features with the number of channels \(3+M \times F'\), which is then used to extract fine-grained features by four shared MLP layers (64, 64, 64, 128). The skip-connection method is employed to connect local signature and these intermediate layers, followed by a shared full-connected layer (1024) and a max pooling operation over feature channels to obtain a global feature for the entire point cloud. We finally apply three shared MLP layers (512, 256, 40) and dropout operation with a keep probability of 0.5 to transform global feature to 40 categories. Besides, the activation function ReLU with batch normalization is used in each layer, and the number of neighbors \(k\) is set to 20.

\paragraph{Training details.} During the training, our optimizer model is Adam \cite{kingma2014adam} with momentum 0.9, and we set batch size 32 and learning rate starts from 0.005  and then is divided by 2 every 20 epochs to 0.00001. The decay rate for batch normalization is initially set to 0.7 and increases to 0.99 gradually. Our model is trained on a NVIDIA GTX1080Ti GPU and TensorFlow v1.6.

\paragraph{Results.} Table~\ref{tab:cls_result} compares our results and complexity with several recent state-of-the-art works, and our model achieves the best performance on the ModelNet40 benchmark, and it outperforms the previous state-of-the-art model \textit{DGCNN} by 0.2\% accuracy. 

To compare the complexity, we measured the model complexity and the computational complexity using the model size and forward time respectively. We also evaluated and listed in Table~\ref{tab:cls_result} the same metrics for all the available models in the same experimental environment. Although \textit{PointNet} achieves the best computational complexity, our model outperforms it by 3.1\% accuracy, which leads to the fact that our model achieves the best trade-off between accuracy and complexity.

\begin{table}[h]
  \caption{Classification results on ModelNet40 dataset.}
  \label{tab:cls_result} \centering
  \begin{tabular}{ccccc}
    \toprule[0.96pt]
      &  \begin{tabular}[c]{@{}c@{}}MEAN CLASS \\ ACCURACY (\%) \end{tabular} & \begin{tabular}[c]{@{}c@{}}OVERALL \\ ACCURACY  (\%) \end{tabular} & \begin{tabular}[c]{@{}c@{}}MODEL \\ SIZE (MB) \end{tabular} & \begin{tabular}[c]{@{}c@{}}FORWARD \\ TIME (MS) \end{tabular}\\
    \midrule
    VOXNET \cite{maturana2015voxnet}                 & 83.0                                                           & 85.9                              & -                          & -  \\
    POINTNET  \cite{qi2017pointnet}               & 86.0                                                           & 89.2                                      & 41.8                  & \textbf{14.7}  \\
    POINTNET++  \cite{qi2017pointnet++}           & -                                                              & 90.7                                  & 19.9                  & 32.0      \\
    KD-NET  \cite{klokov2017escape}                & -                                                              & 91.8                                       & -                 & -  \\
    KC-NET  \cite{shen2018mining}               & -                                                              & 91.0                                            & -            & -  \\
    DGCNN  \cite{wang2018dynamic}                  & \textbf{90.2}                                                           & 92.2                   & 22.1             & 52.0                          \\
    \midrule
    OURS                   & 89.7                                                           & \textbf{92.4}                                 & 22.9                     & 27.9    \\ 
    \bottomrule[0.96pt]
  \end{tabular}
\end{table}

\paragraph{Ablation study.} We also test our classification model with different settings on the ModelNet40 benchmark \cite{wu20153d} . In particular, we analyze the effectiveness of the GAPLayer, attention pooling layer, and also different numbers of multiple heads and encoding channels. 

Table~\ref{tab:ablation_1} represents the advantages of our GAPLayer and attention pooling layer. It shows that attention pooling layer leads to 0.6\% accuracy. Constant-GAPLayer indicates a model with the same structure as our GAPLayer, but all the coefficients are set to equal constants, and it indicates the effectiveness of graph attention mechanism and our GAPLayer model that  leads to 0.7\% accuracy. 

For what concerns the impact of different numbers of heads \(M\) and encoding channels \(F'\). Table~\ref{tab:ablation_2} indicates that appropriate numbers are beneficial to local feature extraction, however the performance degenerates when the numbers become further larger.

\begin{table*}[!]\footnotesize
  \caption{Semantic part segmentation results on ShapeNet part dataset.}
  \label{tab:seg_result} \centering
 \begin{tabular}{p{1.2cm}|p{0.4cm}|p{0.3cm}p{0.3cm}p{0.3cm}p{0.3cm}p{0.3cm}p{0.3cm}p{0.3cm}p{0.3cm}p{0.3cm}p{0.3cm}p{0.3cm}p{0.3cm}p{0.3cm}p{0.3cm}p{0.3cm}p{0.3cm}}
    \toprule[0.5pt]
    & avg & air. & bag  & cap  & car  & cha. & ear. & gui. & kni. & lam. & lap. & mot. & mug  & pis. & roc. & ska. & tab. \\ \midrule
\begin{tabular}[c]{@{}c@{}}shapes\\ number\end{tabular} &      & 2690 & 76   & 55   & 898  & 3758  & 69                                                  & 787    & 392   & 1547 & 451    & 202   & 184  & 283    & 66     & 152                                                   & 5271  \\ \midrule
pointnet                                                & 83.7 & 83.4 & 78.7 & 82.5 & 74.9 & 89.6  & 73.0                                                & 91.5   & 85.9  & 80.8 & 95.3   & 65.2  & 93.0 & 81.2   & 57.9   & 72.8                                                  & 80.6  \\
pointnet++                                              & \textbf{85.1} & 82.4 & 79.0 & 87.7 & 77.3 & 90.8  & 71.8                                                & 91.0   & 85.9  & \textbf{83.7} & 95.3   & \textbf{71.6}  & 94.1 & 81.3   & 58.7   & \textbf{76.4}                                                  & \textbf{82.6}  \\
kd-net                                                  & 82.3 & 82.3 & 74.6 & 74.3 & 70.3 & 88.6  & 73.5                                                & 90.2   & 87.2  & 81.0 & 94.9   & 57.4  & 86.7 & 78.1   & 51.8   & 69.9                                                  & 80.3  \\
dgcnn                                                   & \textbf{85.1} & \textbf{84.2} & 83.7 & 84.4 & 77.1 & \textbf{90.9}  & \textbf{78.5}                                                & \textbf{91.5}   & \textbf{87.3}  & 82.9 & 96.0   & 67.8  & 93.3 & \textbf{82.6}   & 59.7   & 75.5                                                  & 82.0  \\ \midrule
ours                                                    & 84.7 & \textbf{84.2} &\textbf{84.1} & \textbf{88.8} & \textbf{78.1} & 90.7  & 70.1                                                & 91.0   & \textbf{87.3}  & 83.1 & \textbf{96.2}   & 65.9  & \textbf{95.0} & 81.7   & \textbf{60.7}   & 74.9                                                  & 80.8  \\ 
   \bottomrule[0.5pt]
  \end{tabular}
\end{table*}

\begin{table}[!]
\begin{floatrow}
\ttabbox{\caption{Effectiveness of different components.}}
{\label{tab:ablation_1} \centering
\begin{tabular}{cccc}
\toprule[0.96pt]
Components & \begin{tabular}[c]{@{}c@{}}Overall\\ Accuracy (\%)\end{tabular} \\
\midrule
\begin{tabular}[c]{@{}c@{}}Constant-GAPLayer\\  + Attention Pooling  \end{tabular}   & 91.7                                                       \\
GAPLayer                     & 91.8                                                       \\
GAPLayer + Attention Pooling                 & 92.4                                                       \\
\bottomrule[0.96pt]
\end{tabular}}

\ttabbox{\caption{Effectiveness of different numbers of heads and encoding channels.}}
{\label{tab:ablation_2} \centering
\begin{tabular}{cccc}
\toprule[0.96pt]
Heads & \begin{tabular}[c]{@{}c@{}}Encoding\\  Channels \(F'\)\end{tabular} & \begin{tabular}[c]{@{}c@{}}Overall\\ Accuracy  (\%)\end{tabular} \\
\midrule
1         & 8                         & 91.6                                                       \\
4          & 8                            & 91.9                                                       \\
4          & 16                           & 92.4                                                       \\
8         & 16                       & 91.7                                                       \\

\bottomrule[0.96pt]
\end{tabular}}
\end{floatrow}
\end{table}

\subsection{Semantic part segmentation}
\paragraph{Dataset.} We evaluate our segmentation model on ShapeNet part dataset \cite{yi2016scalable} in semantic part segmentation task that is to classify part category for each point from a mesh model. The dataset consists of 16,881 CAD shapes of 16 categories, and each point from a model is annotated with a class of 50 part classes. Besides, each shape model is labeled with several but less than 6 parts. We follow the same sampling strategy as Section~\ref{cls_point} to sample 2,048 points uniformly, and split dataset into 9,843 models for training and 2,468 models for testing in our experiment.

\paragraph{Model structure.} Our segmentation model shown in Figure~\ref{fig:model_structure}  (bottom branch) is to predict a part category label for each point in the point cloud. We firstly use the same spatial transformer network and the GAPLayer as Section~\ref{cls_point}, followed by shared MLP layers (64, 64, 128). Then the second GAPLayer with 4 heads and 128 encoding channels is applied, followed by shared MLP layers (128, 128, 512) to obtain representations with 512 channels, which are concatenated with local signature generated from corresponding attention pooling layer of GAPLayer. The aggregated feature applies a shared full-connected layer (1024) and a max pooling to obtain a global feature, which is then duplicated 2048 times and finally applies four shared full-connected layers (256,256,128,50) with dropout probability 0.6 to transform the global feature to 50 part categories.

\paragraph{Training details.} The training setting is similar to the setting in classification task, except that batch size is set to 8, number of neighbors \(k\) is set to30,  and we distribute the task to two NVIDIA TESLA V100 GPUs.

\paragraph{Results.} We use the mean Intersection over Union (mIoU) \cite{qi2017pointnet} as our evaluation scheme to align the evaluation metric. The IoU of each shape is calculated by averaging  IoUs for all parts that fall into the same category, then the mIoU is the mean IoUs for all shapes from testing dataset.

\begin{figure*}[t]
  \centering
   \subfigure[Visualization]{\label{fig:objects}\includegraphics[width=0.45\linewidth]{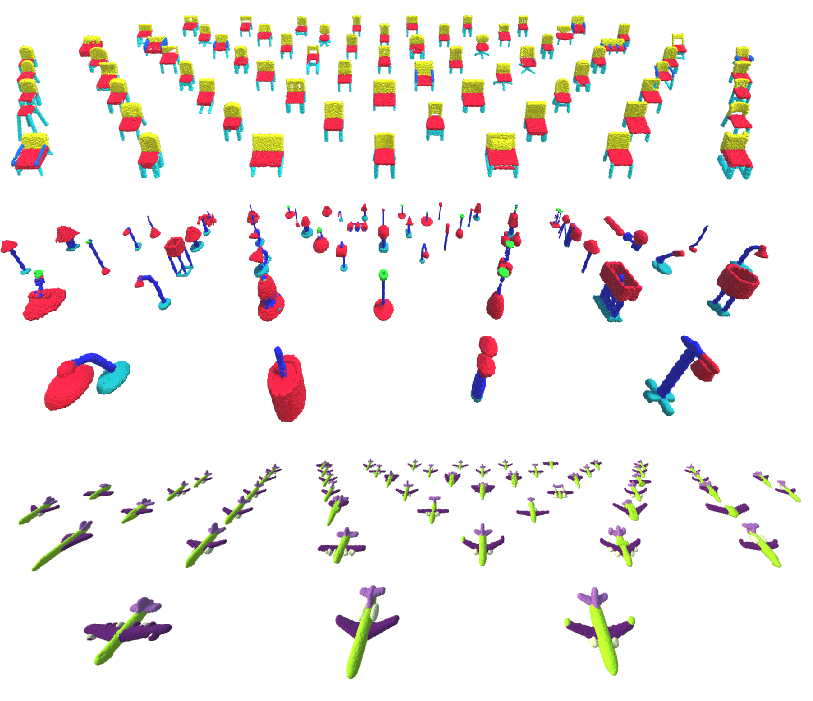}}
   \subfigure[Comparison]{\label{fig:compare}\includegraphics[width=0.34\linewidth]{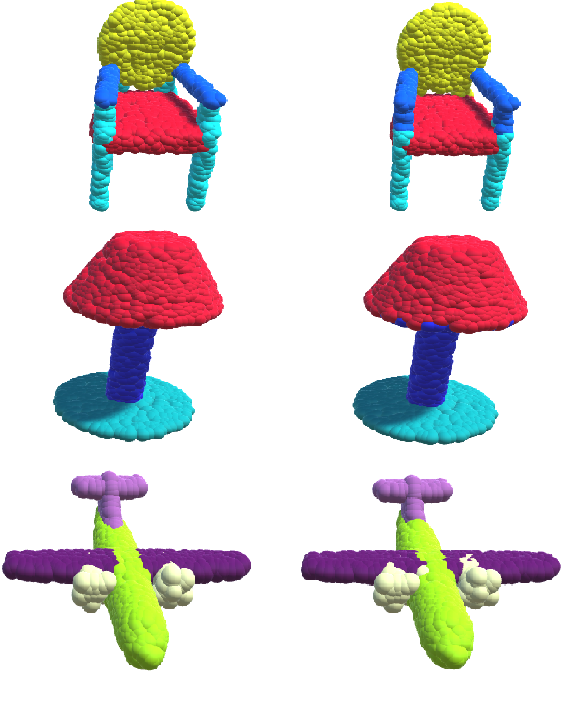}}
  \caption{Visualization of semantic part segmentation results. Figure \subref{fig:objects} visualizes some samples: chair (top), lamp (middle),  and airplane (bottom). While Figure \subref{fig:compare} visualizes the difference between ground truth (left) and prediction (right).}
  \label{fig:visual}
 \end{figure*}

Table~\ref{tab:seg_result} shows that our model achieves competitive results on the ShapeNet part dataset \cite{yi2016scalable}. Our model wins 8 categories for part segmentation compared with 6 winning categories from \textit{DGCNN} \cite{wang2018dynamic}, although it outperforms ours by 0.4\% accuracy. Figure~\ref{fig:objects} represents some shapes from our results, we also visualize the difference between ground truth and our prediction results as shown in  Figure~\ref{fig:compare} , where left shapes indicate ground truth and right shapes show our prediction results.

\section{Conclusions}
In this paper, we propose a graph attention based point neural network, named GAPNet, to learn shape representations for point cloud. Experiments show state-of-the-art performance in shape classification and semantic part segmentation tasks. The success of our model also verifies the fact that graph attention network shows efficiency in not only similarity computation for graph nodes, but also geometric relationship understanding.

In the future, we can further explore several research avenues. For example, some applications, such as autonomous vehicle, normally need to process very large-scale point cloud data. As a result, how to efficiently and robustly deal with large-scale data would be a worthwhile work. Furthermore, it would be interesting to develop an efficient \textit{CNN}-like operation for unstructured data analysis.

\subsubsection*{Acknowledgments}
The HumanDrive project is a CCAV - Innovate UK funded R\&D project (Project ref: 103283).

%

\end{document}